\documentclass[conference]{IEEEtran}
\IEEEoverridecommandlockouts
\usepackage{cite}
\usepackage{graphicx}
\usepackage{epstopdf}
\usepackage{amsmath,amssymb,amsfonts}
\usepackage{algorithmic}
\usepackage{textcomp}
\usepackage{xcolor}
\usepackage[ruled,lined,linesnumbered]{algorithm2e}
\usepackage{makecell}
\usepackage{extarrows}
\usepackage{multicol}
\usepackage{subcaption}
\usepackage{multirow}
\usepackage{bm}
\usepackage{colortbl}
\usepackage{amsthm}
\usepackage{mathrsfs}
\usepackage{booktabs}
\usepackage{siunitx}
\usepackage{tabularx,booktabs}
\usepackage[flushleft]{threeparttable}
\usepackage[short]{optidef}

\usepackage{xcolor}
\usepackage{xparse}
\usepackage[normalem]{ulem}
\usepackage[most]{tcolorbox}
\definecolor{promptframe}{HTML}{2C3E50}
\definecolor{promptback}{HTML}{F4F6F8}
\definecolor{promptslot}{HTML}{B22222}
\newtcolorbox{promptbox}[1][]{enhanced, breakable, sharp corners,
  colback=promptback, colframe=promptframe, boxrule=0.4pt, left=6pt, right=6pt,
  top=4pt, bottom=4pt, fonttitle=\bfseries\sffamily\small,
  coltitle=white, colbacktitle=promptframe,
  attach boxed title to top left={xshift=6pt, yshift=-2pt},
  boxed title style={sharp corners, boxrule=0pt, top=1pt, bottom=1pt,
    left=4pt, right=4pt}, #1}

\newif\ifshownotes
\shownotesfalse     

\newcommand{\defineauthor}[3]{%
  \expandafter\gdef\csname author@#1@name\endcsname{#2}%
  \expandafter\gdef\csname author@#1@color\endcsname{#3}%
}

\NewDocumentCommand{\note}{m m}{%
  \ifshownotes
    {\textbf{[\csname author@#1@name\endcsname: #2]}}
  \fi
}

\defineauthor{GLM5}{GLM-5}{purple}
\defineauthor{SK}{Codex}{blue}
\defineauthor{SP}{Supervisor}{red}

\definecolor{proposedblue}{RGB}{242,242,255}     
\definecolor{proposedblueB}{RGB}{255, 220, 225}   

\def\BibTeX{{\rm B\kern-.05em{\sc i\kern-.025em b}\kern-.08em
T\kern-.1667em\lower.7ex\hbox{E}\kern-.125emX}}

\begin{document}

\title{CoVSpec: Efficient Device--Edge Co-Inference for Vision-Language Models via Speculative Decoding
}

\author{

\IEEEauthorblockN{
Yuanyuan Jia\IEEEauthorrefmark{2}\IEEEauthorrefmark{3},  
Shunpu Tang\IEEEauthorrefmark{2}\IEEEauthorrefmark{3}, 
Qianqian Yang\IEEEauthorrefmark{2}
}

\IEEEauthorblockA{ 
\IEEEauthorrefmark{2}College of Information Science and Electronic Engineering, Zhejiang University, Hangzhou, China \\
Email: \{labulado, tangshunpu, qianqianyang20\}@zju.edu.cn
}

\thanks{
This work is partly supported by the National Key R\&D Program of China under Grant 2024YFE0200802, by the NSFC under grant No. 62293481 and No. 62571487,  and by the Zhejiang Provincial Natural Science Foundation of China under Grant No. LZ25F010001.  (Corresponding author: Q. Yang.)
}
\thanks{\IEEEauthorrefmark{3}Equal contribution. }
}

\maketitle

\begin{abstract}
Vision-language models (VLMs) have demonstrated strong capabilities in multimodal perception and reasoning. However, deploying large VLMs on mobile devices remains challenging due to their substantial computational and memory demands. A practical alternative is device--edge co-inference, where a lightweight draft VLM on the mobile device collaborates with a larger target VLM on the edge server via speculative decoding. Nevertheless, directly extending speculative decoding to VLMs suffers from severe inefficiency due to excessive visual-token computation and high communication overhead. To address these challenges, we propose CoVSpec, an efficient collaborative speculative decoding framework for VLM inference. Specifically, we first develop a training-free visual token reduction framework that prunes redundant visual tokens on the mobile device by jointly considering query relevance, token activity, and low-rank dependency. Moreover, we design an adaptive drafting strategy that dynamically adjusts both the verification frequency and the draft length. In addition, we introduce a parallel branching mechanism with decoupled verification--correction to improve draft-side utilization during target-side verification and reduce correction-related transmission overhead. Experiments on multiple benchmarks show that CoVSpec achieves up to 2.21$\times$ higher throughput than target-only inference and reduces communication overhead by more than 96\% compared with baselines, without compromising task accuracy.

\end{abstract}
\begin{IEEEkeywords}
Vision-language models, collaborative inference, speculative decoding, edge intelligence, and visual token reduction.
\end{IEEEkeywords}

\section{Introduction}
Recently, vision-language models (VLMs) have demonstrated strong capabilities in multimodal perception and reasoning, achieving substantial gains across a wide range of applications~\cite{zhang2024vision}. However, practical VLM deployment faces a trade-off between model capability and deployment cost. Specifically, small VLMs can run on mobile devices due to their lower computation and memory demands, yet they often lack the reasoning capability needed for complex tasks. In contrast, large VLMs provide significantly stronger performance, but their high computation and memory costs typically restrict deployment to remote servers. As a result, using only a small on-device VLM may lead to degraded performance, while fully offloading inference to the server side may incur high latency and place a heavy load on the remote server.

Motivated by the recent concepts of edge intelligence~\cite{Edge_AI} and AI-RAN~\cite{11474793}, which bring AI capabilities toward the network edge, a practical solution is device--edge co-inference, where a lightweight model on the mobile device performs low-cost drafting, while a large model at a nearby edge server verifies and corrects the drafted tokens when needed. To implement this idea, the authors in~\cite{JP_hybrid} first proposed a hybrid inference framework between an on-device small language model and a remote large language model based on speculative decoding, and designed an uncertainty-aware mechanism to skip unnecessary verification, thereby reducing communication cost and improving throughput. Following this work, the authors in~\cite{ning2025dssd} further improved the system efficiency by optimizing transmission and verification strategies.

However, directly extending these works to VLM inference in mobile edge networks remains inefficient. Specifically, the large number of visual tokens in VLMs leads to heavy mobile-side cost, which is especially problematic for practical systems with computation- and memory-limited mobile devices. Moreover, their efficiency is further limited by frequent device--edge verification and correction, which introduces substantial communication overhead over wireless links. As a result, directly applying speculative decoding to such device--edge co-inference systems may fail to improve end-to-end efficiency and may even underperform server-only target-model inference in some cases. Although recent speculative decoding methods~\cite{cai2024medusa,li2024eagle,ji2025specvlm} and visual token reduction frameworks~\cite{11092998,guo2025crop,alvar2025divprune} have been studied for VLM acceleration, most of them target standalone or server-side inference rather than practical mobile deployment, and therefore do not provide a systematic design that jointly considers on-device drafting cost, device--edge verification frequency, and communication overhead.

To address these issues, we propose CoVSpec, an efficient device--edge co-inference framework for VLM inference. Specifically, our key idea is to let the mobile-side draft VLM operate on a compact visual token set, while the edge-server target VLM still performs verification with the full visual tokens, so that on-device drafting cost is reduced without compromising verification accuracy. The main contributions of this work are summarized as follows.
\begin{itemize}
    \item We propose a training-free visual token selection framework for efficient mobile-side drafting, which selects compact and informative visual tokens by jointly considering text-conditioned relevance, token activity, and low-rank redundancy.
    \item We introduce a communication-aware adaptive drafting strategy with margin gating, parallel branching, and decoupled verification-correction, which jointly controls the edge-server verification frequency, adapts the draft length to channel conditions, and reduces transmission overhead.
    \item We conduct extensive experiments on multiple benchmarks. In particular, CoVSpec achieves up to $2.21\times$ higher decoding speed than the target-only baseline and reduces communication overhead by over $96\%$ compared with conventional device--edge speculative decoding, while maintaining competitive answer accuracy.
\end{itemize}

\section{System Model}
In this section, we present the system model of the proposed device--edge collaborative VLM inference framework, where a lightweight draft model is deployed on the mobile device while a large target model is hosted at the edge. We first describe the co-inference pipeline, and then introduce the transmission model.
\begin{figure*}[t]
    \centering
    \includegraphics[width=0.9\textwidth]{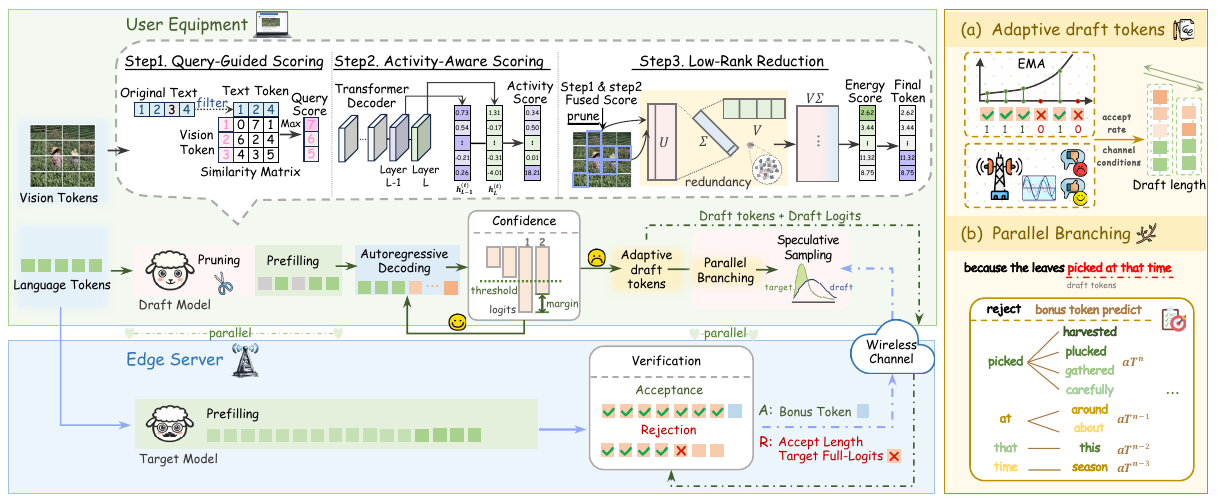}
    \caption{Illustration of the proposed CoVSpec, a communication-aware device-edge collaborative speculative decoding framework for VLM inference.}
    \label{fig:pipeline}
\end{figure*}
\subsection{Collaborative VLM Inference Pipeline}
We consider a mobile device that receives a visual input $\bm{x}$ together with a text query $\bm{q}$. Let $\bm{V}=\{v_1,v_2,\ldots,v_N\}$ denote the full visual token set extracted from $\bm{x}$, and let $\mathcal{W}$ denote the output vocabulary. To reduce the computational overhead on the mobile device, only a reduced visual context $\hat{\bm{V}} \subseteq \bm{V}$ is selected for drafting, whereas the edge-side target model always uses the full visual token set $\bm{V}$ for verification.

Specifically, in each round, starting from the current decoding position $t$, the device-side draft VLM $f_d(\cdot)$ autoregressively generates $k$ consecutive candidate tokens $(\hat y_{t},\hat y_{t+1},\ldots,\hat y_{t+k-1})$. Let $\tau\in\{t,t+1,\ldots,t+k-1\}$ index an arbitrary position within these $k$ draft tokens. At position $\tau$, the draft VLM $f_d(\cdot)$ takes the selected visual token set $\hat{\bm{V}}$, the text query $\bm{q}$, and the previously generated token prefix $\bm{y}_{<\tau}$ as input, and produces the logits over the output vocabulary, which can be expressed as
\begin{equation}
\bm{z}_d^{(\tau)}=f_d(\hat{\bm{V}},\bm{q},\bm{y}_{<\tau}),
\end{equation}
where $\bm{z}_d^{(\tau)}\in\mathbb{R}^{|\mathcal{W}|}$ denotes the logit vector produced by the device-side VLM, and $|\mathcal{W}|$ denotes the vocabulary size. The device-side next-token distribution is then obtained by applying the softmax operation, given by
\begin{equation}
p_d(w|\tau)=\frac{\exp(z_{d,w}^{(\tau)})}{\sum_{u\in\mathcal{W}}\exp(z_{d,u}^{(\tau)})},\quad w\in\mathcal{W},
\end{equation}
where $z_{d,w}^{(\tau)}$ denotes the $w$-th entry of $\bm{z}_d^{(\tau)}$, and $p_d(w|\tau)$ denotes the probability that the next token equals $w$ under the device-side draft model. The device then samples a draft token $\hat y_\tau\sim p_d(\cdot|\tau)$ and moves on to the next position. Once all $k$ draft tokens are produced, they are packed together with their corresponding draft logits, and transmitted to the edge for verification.

On the edge side, benefiting from the transformer's prefill process that computes logits for all drafted positions simultaneously, the target VLM $f_t(\cdot)$ verifies all $k$ drafted tokens in parallel using the full visual token set. Specifically, for each draft position $\tau$, the target model produces
\begin{equation}
\bm{z}_t^{(\tau)}=f_t(\bm{V},\bm{q},\bm{y}_{<\tau}),
\end{equation}
where $\bm{z}_t^{(\tau)}\in\mathbb{R}^{|\mathcal{W}|}$ denotes the target-side logit vector at position $\tau$. The corresponding target-side token distribution is then given by
\begin{equation}
p_t(w| \tau)=\frac{\exp(z_{t,w}^{(\tau)})}{\sum_{u\in\mathcal{W}}\exp(z_{t,u}^{(\tau)})},\quad w\in\mathcal{W}.
\end{equation}

Following the standard speculative decoding, the edge checks each draft position and accepts the drafted token $\hat y_\tau$ with probability
\begin{equation}
\alpha_\tau=\min\left(1,\frac{p_t(\hat y_\tau| \tau)}{p_d(\hat y_\tau| \tau)}\right).
\end{equation}
If accepted, $\hat y_\tau$ is directly appended to the decoding prefix. Otherwise, once the first rejection occurs, a correction is drawn from the residual distribution at that position, given by
\begin{equation}
r_{\tau}(w)=
\frac{\max\!\big(p_t(w| \tau)-p_d(w| \tau),\,0\big)}
{\sum_{u\in\mathcal{W}}\max\!\big(p_t(u| \tau)-p_d(u| \tau),\,0\big)},
\quad w\in\mathcal{W},
\label{eq:residual_sampling}
\end{equation}
where $r_{\tau}(w)$ denotes the residual distribution used to resample the corrected token. Building on this pipeline, we further introduce several improvements, including visual token reduction, adaptive drafting, and parallel branching, which will be detailed in the next section.

\subsection{Transmission Model}

In each interaction round, the device and the edge exchange one uplink message and one downlink message. In the uplink, the device uploads the drafted token IDs together with only the draft logits associated with the drafted tokens, rather than the full-vocabulary draft distribution. In the downlink, if all drafted tokens are accepted, the edge returns one bonus token; otherwise, it returns the rejection position together with the full-vocabulary target logits for device-side correction.

Let $S_{\mathrm{up}}$ and $S_{\mathrm{down}}$ denote the uplink and downlink payload sizes, respectively, measured in bits. The uplink payload is given by
\begin{equation}
S_{\mathrm{up}}=
N_{\mathrm{draft}}\,b_{\mathrm{id}}+
N_{\mathrm{logit}}^{\mathrm{draft}}\,b_{\mathrm{logit}},
\end{equation}
where $N_{\mathrm{draft}}$ is the number of drafted tokens, $b_{\mathrm{id}}$ is the number of bits used to represent each token ID, $N_{\mathrm{logit}}^{\mathrm{draft}}$ is the number of uploaded draft logits, and $b_{\mathrm{logit}}$ is the number of bits used for each draft logit. Likewise, the downlink payload is given by
\begin{equation}
S_{\mathrm{down}}=
\begin{cases}
b_{\mathrm{acc}}+b_{\mathrm{bonus}}, & \text{if all drafted tokens accepted},\\
b_{\mathrm{rej}}+|\mathcal{W}|\,b_{\mathrm{logit}}^{\mathrm{tar}}, & \text{otherwise},
\end{cases}
\end{equation}
where $b_{\mathrm{acc}}$ denotes the number of bits required to encode the accepted length, $b_{\mathrm{bonus}}$ denotes the number of bits for the bonus token, $b_{\mathrm{rej}}$ denotes the number of bits used to indicate the rejection position, and $b_{\mathrm{logit}}^{\mathrm{tar}}$ denotes the number of bits for each target logit. In this work, both draft and target logits are quantized in float16 format for transmission.

To characterize the communication cost, we adopt a point-to-point wireless link model with bandwidth $B$ and received signal-to-noise ratio $\mathrm{SNR}$ in dB, in which the communication latency can be expressed as
\begin{equation}
T_{\mathrm{comm}}
=
\frac{S_{\mathrm{up}}+S_{\mathrm{down}}}{B\log_2(1+10^{\mathrm{SNR}/10})}.
\end{equation}

\section{Proposed Framework}

In this section, we present the proposed CoVSpec framework as shown in Fig. \ref{fig:pipeline}. Specifically, we first introduce the proposed visual token selection strategy, then describe the adaptive drafting mechanism, and finally present the parallel branching scheme with decoupled verification-correction.
\subsection{Visual Token Selection}

To reduce the visual token burden on the device-side draft model, we perform visual token selection only for drafting, while the edge-side target model always verifies with the full visual tokens. Specifically, we first score each visual token by jointly considering its query relevance and token activity, and then further compress the resulting candidate set via low-rank redundancy suppression to obtain the compact visual token set.

\subsubsection{Query-Aware Scoring}
To measure text-conditioned relevance, we first extract a keyword subset $\mathcal{Q}$ from the input query $\bm{q}$ using spaCy, which removes less informative function words and keeps the key semantic terms. Let $e(v_i)$ and $e(q_j)$ denote the embeddings of the $i$-th visual token and the $j$-th query element, respectively. 
The query-aware score of the $i$-th visual token is then defined as
\begin{equation}
S_i^{\mathrm{query}}
=
\max_{q_j\in\mathcal{Q}}
\frac{\langle e(v_i), e(q_j)\rangle}
{\|e(v_i)\|_2\,\|e(q_j)\|_2},
\quad i=1,\ldots,N,
\end{equation}
where $\langle \cdot,\cdot\rangle$ denotes the inner product, and $S_i^{\mathrm{query}}$ measures the maximum cosine similarity between visual token $v_i$ and the query keywords in $\mathcal{Q}$.
\subsubsection{Token Activity Scoring}
While query-aware scoring captures text-conditioned relevance, it may miss visual tokens that are important for prediction but are not explicitly aligned with the query keywords. To complement this, we further measure how actively each visual token evolves across the late prefill layers of the target model. Empirically, we observe that visual tokens with larger representation changes in the final prefill layers tend to have a stronger impact on the model prediction. Motivated by this observation, we use the inter-layer variation magnitude of each visual token as a proxy for its importance. Specifically, the activity score of token $v_i$ is defined as
\begin{equation}
S_i^{\mathrm{act}}
=
\frac{1}{K}
\sum_{\ell=L-K+1}^{L}
\left\|
h_i^{(\ell)}-h_i^{(\ell-1)}
\right\|_2,
\quad i=1,\ldots,N,
\end{equation}
where $h_i^{(\ell)}$ denotes the hidden representation of token $v_i$ at layer $\ell$, $L$ is the total number of prefill layers, and $K$ is the number of late layers used for activity measurement. Finally, we normalize the two scores to the same scale and combine them
as
\begin{equation}
S_i
=
\lambda \bar S_i^{\mathrm{query}}
+
(1-\lambda)\bar S_i^{\mathrm{act}},
\quad i=1,\ldots,N,
\end{equation}
where $\lambda\in[0,1]$ balances query relevance and token activity.
\subsubsection{Low-Rank Redundancy Reduction}
Based on $\{S_i\}_{i=1}^N$, a straightforward approach is to retain the $M$ highest-scoring visual tokens, denoted by $\bm{V}_M$. However, these tokens may still contain overlapping visual information. To further reduce such redundancy, we extract a compact set of representative tokens via low-rank reduction. Specifically, let $z_i\in\mathbb{R}^{d}$ denote the representation of the $i$-th preselected token, and stack them into a matrix
\begin{equation}
\bm{Z}_M=
\begin{bmatrix}
z_1, z_2, \cdots, z_M
\end{bmatrix}
\in\mathbb{R}^{d\times M}.
\end{equation}
We then perform truncated SVD on $\bm{Z}_M$, given by
\begin{equation}
\bm{Z}_M \approx \bm{U}_r\bm{\Sigma}_r\bm{V}_r^\top,
\end{equation}
where $r<\min(M,d)$ is the retained rank. The energy of each preselected token in the reduced subspace is measured as
\begin{equation}
c_i=
\left\|
\left(\bm{V}_r\bm{\Sigma}_r\right)_{i,:}
\right\|_2^2,
\quad i=1,\ldots,M.
\end{equation}
Finally, according to $\{c_i\}_{i=1}^M$, we retain the $B_{\mathrm{vis}}$ tokens with the largest subspace energies to form the reduced visual context $\hat{\bm{V}}$, where $B_{\mathrm{vis}}<M\le N$.
\subsection{Adaptive Drafting}
To reduce unnecessary device--edge interactions while adapting to the current channel condition, we employ two mechanisms: a margin-based gating rule to skip verification for high-confidence draft tokens, and an adaptive strategy that adjusts the draft length online.
\subsubsection{Margin-Based Gating}
At decoding step $\tau$, the device uses the probability margin between the top-1 and top-2 candidates as an uncertainty indicator for the current draft token. Specifically, let $p^{(\tau)}_{(1)}$ and
$p^{(\tau)}_{(2)}$ denote the largest and second-largest probabilities in the draft distribution $p_d(\cdot|\tau)$, respectively. The margin is then defined as
\begin{equation}
m_{\tau} = p^{(\tau)}_{(1)} - p^{(\tau)}_{(2)},
\end{equation}
where a larger margin indicates higher confidence in the top-1 prediction and thus lower uncertainty.  Motivated by this observation, we define the gating decision as
\begin{equation}
g_{\tau}=
\begin{cases}
0, & m_{\tau} \ge \gamma,\\
1, & m_{\tau} < \gamma,
\end{cases}
\end{equation}
where $\gamma$ is a predefined margin threshold.  If $m_{\tau} \ge \gamma$, the current token is regarded as sufficiently confident and is directly committed, so decoding continues locally without edge verification. Otherwise, a new verification round is triggered.
\subsubsection{Adaptive Drafting Length}
Rather than keeping the draft length fixed, we adjust it online based on the recent acceptance behavior of draft tokens and the transmission latency that a rejection would incur.

To measure the recent acceptance condition, we maintain an exponential
moving average (EMA) of the token-level acceptance probability. Let
$\hat p_{\tau}$ denote the estimate at round $\tau$, initialized as
$\hat p_0=1.0$. In round $\tau$, the device proposes $k_{\tau}$
speculative tokens, among which $n_{\mathrm{acc}}$ are accepted by the
verifier. We model these token-level outcomes as Bernoulli observations, where an accepted token corresponds
to $a=1$ and a rejected token corresponds to $a=0$. The estimate is then updated as
\begin{equation}
\hat p_{\tau} \leftarrow (1-\eta)\hat p_{\tau} + \eta a,\qquad a\in\{0,1\},
\end{equation}
where $\eta\in(0,1)$ is the smoothing factor. Accordingly, in round $\tau$, the EMA is updated $n_{\mathrm{acc}}$ times with $a=1$ and $k_{\tau}-n_{\mathrm{acc}}$ times with $a=0$. In this way, $\hat p_{\tau}$ reflects how likely a speculative token is to be accepted under the current condition.

Moreover, we estimate the transmission latency under the rejection case as
\begin{equation}
T_{\mathrm{comm},\tau}^{\mathrm{rej}}
=
\frac{S_{\mathrm{up}}^{\mathrm{rej}}+S_{\mathrm{down}}^{\mathrm{rej}}}
{B\log_2(1+10^{\mathrm{SNR}/10})},
\end{equation}
where $S_{\mathrm{up}}^{\mathrm{rej}}$ and
$S_{\mathrm{down}}^{\mathrm{rej}}$ denote the uplink and downlink payload
sizes when a rejection occurs.

\textcolor{black}{
Using these two quantities, we adjust the draft length as follows. When
the recent acceptance drops, we decrease the draft length to avoid unnecessary cost. When the
recent acceptance is reliable, we further check the communication latency:
if the latency is large, we keep the draft length unchanged; otherwise,
we increase it. To make this hierarchical logic explicit, we define
\begin{equation}
\phi_{\tau}=
\begin{cases}
-1, & \hat p_{\tau} \le p_{\mathrm{low}},\\
1, & \hat p_{\tau} \ge p_{\mathrm{up}},\ T_{\mathrm{comm},\tau}^{\mathrm{rej}}\le T_{\mathrm{ref}},\\
0, & \text{otherwise},
\end{cases}
\end{equation}
and update the draft length as
\begin{equation}
k_{\tau+1}=\operatorname{round}\!\left(k_{\tau}s^{\phi_{\tau}}\right),
\end{equation}
where $p_{\mathrm{up}}$ and $p_{\mathrm{low}}$ are the acceptance
thresholds, $T_{\mathrm{ref}}$ is the reference communication-latency
threshold, and $s>1$ is the scaling factor. 
} Finally, the updated draft length is clipped to the valid range:
\begin{equation}
k_{\tau+1}=\min \bigg(k_{\max},\max(k_{\min},k_{\tau+1})\bigg).
\end{equation}

\subsection{Parallel Branching with Decoupled Verification-Correction}

To reduce both the exposed draft-side latency and the correction-related transmission within each verification round, we combine parallel branching with a decoupled verification-correction protocol. 
\begin{table*}[t]
\centering
\small
\setlength{\tabcolsep}{3.5pt}
\renewcommand{\arraystretch}{1.12}
\begin{tabular}{lcccccccccccc}
\toprule
\multirow{2}{*}{framework}
& \multicolumn{4}{c}{VQAv2}
& \multicolumn{4}{c}{MMMU}
& \multicolumn{4}{c}{MMBench} \\
\cmidrule(lr){2-5}
\cmidrule(lr){6-9}
\cmidrule(lr){10-13}
& Spd. $\uparrow$ & Acc. $\uparrow$ & Comm. $\downarrow$ & Cost Red. $\uparrow$
& Spd. $\uparrow$ & Acc. $\uparrow$ & Comm. $\downarrow$ & Cost Red. $\uparrow$
& Spd. $\uparrow$ & Acc. $\uparrow$ & Comm. $\downarrow$ & Cost Red. $\uparrow$ \\
\midrule

Edge-only
& 1.00$\times$ & \textbf{74.70} & -- & 0.00
& \underline{1.00$\times$} & \textbf{58.30} & -- & 0.00
& \underline{1.00$\times$} & \underline{88.00} & -- & 0.00 \\
Device-only
& 3.41$\times$ & 56.20 & -- & --
& 3.44$\times$ & 40.00 & -- & --
& 3.75$\times$ & 72.00 & -- & -- \\

Vanilla SD
& 0.54$\times$ & \textbf{74.70} & 568.09 & -7.32
& 0.54$\times$ & \textbf{58.30} & 599.23 & -10.39
& 0.48$\times$ & \underline{88.00} & 534.51 & -9.74 \\

U-HLM
& \underline{1.13$\times$} & 72.60 & 155.27 & \underline{5.06}
& 0.85$\times$ & 39.60 & 248.51 & \underline{7.85}
& 0.87$\times$ & 82.00 & 181.09 & \underline{7.39} \\

\rowcolor{proposedblue}
\textbf{CoVSpec}
& \textbf{2.21$\times$} & \underline{74.40} & \textbf{16.49} & \textbf{29.75}
& \textbf{1.63$\times$} & \underline{52.17} & \textbf{21.23} & \textbf{15.30}
& \textbf{1.85$\times$} & \textbf{90.00} & \textbf{18.30} & \textbf{22.57} \\
\bottomrule
\end{tabular}
\caption{Performance comparison on VQAv2, MMMU, and MMBench at $\mathrm{SNR}=10$ dB.
Spd., Acc., Comm., and Cost Red. denote the speedup ratio, accuracy, communication cost, and cost reduction percentage, respectively. 
The best and second-best results are highlighted in \textbf{bold} and \underline{underlined}, respectively.}
\label{tab:main_results}
\end{table*}
\subsubsection{Parallel Branching}
In conventional speculative decoding, the next drafting round must wait until the current verification is completed, leaving the draft model idle during edge-side verification. To reduce this idle time, we let the draft model pre-compute possible continuations while the edge is verifying the current draft. Specifically, for a draft segment $\hat y_{\tau},\hat y_{\tau+1},\ldots,\hat y_{\tau+k_{\tau}-1}$, the edge prepares bonus-token branches for different verification outcomes. Let $F_j$ denote the number of bonus-token candidates allocated to position $j$. We adopt a geometrically decaying fan-out, given by
\begin{equation}
F_j=\left\lceil F_0 \rho^j \right\rceil, \qquad
j=0,1,\ldots,k_{\tau},\quad 0<\rho<1,
\end{equation}
so that earlier and more likely verification outcomes receive more bonus-token guesses, while later outcomes receive fewer. For each position, we select the top-$F_j$ draft logits as bonus-token candidates, excluding the token already drafted at that position. Once the target-side verification is completed, the device immediately stops further branch prediction and checks whether the returned outcome matches one of the prepared branches. If a match is found, the corresponding branch is committed directly; otherwise, decoding falls back to the  residual correction path. In this way, the prepared branches serve only to reduce the exposed drafting latency, while correctness is still determined by the target-side verification result.

\subsubsection{Decoupled Verification-Correction}
Different from standard speculative decoding, we move the correction sampling step from the edge to the device~\cite{ning2025dssd}. In the standard design, when a rejection occurs, the edge must evaluate the residual distribution
using both the target-side and draft-side distributions, and therefore needs the device to upload the drafted tokens together with their corresponding draft-side logits, which increases communication overhead. In contrast, in our design, the edge only performs verification. Specifically, if all drafted tokens are accepted, the edge returns only the bonus token. Otherwise, it returns the accepted prefix length together with the full-vocabulary target logits
at the first rejected position, and the device completes residual sampling locally using its own draft logits. In this way, draft-side logits are no longer transmitted for correction, and target-side logits are transmitted only for the first rejected position, which substantially reduces the correction-related communication overhead.


\section{Experiments}

\subsection{Experimental Setup}

We evaluate the proposed framework on VQAv2, MMMU, and MMBench. For each dataset, we randomly sample 50 examples and fix each generation to 1024 new tokens. We adopt a same-family VLM pair, where InternVL2.5-4B and InternVL2.5-78B~\cite{chen2024expanding} serve as the draft and target models, respectively. All experiments are implemented with the Hugging Face Transformers framework. Specifically, the device-side draft model runs on a MacBook Pro equipped with an M5 chip, while the target model is deployed in INT8-quantized form on a server equipped with two NVIDIA RTX 4090 GPUs. The wireless bandwidth is set to $B=5$ MHz, and the SNR is set to $10$ dB. The full visual input consists of 768 visual tokens.

We adopt four metrics to evaluate efficiency, task performance, and inference cost:
1) \emph{Speedup (Spd.)}, the tokens-per-second (TPS) ratio with respect to the edge-only autoregressive baseline;
2) \emph{Communication Cost (Comm.)}, the total transmitted data during device--edge interaction, measured in MB;
3) \emph{Accuracy (Acc.)}, the answer accuracy on each benchmark;
4) \emph{Cost Reduction (Cost Red.)}, the percentage reduction in API cost relative to the edge-only large-model baseline. We note that since the draft model is deployed locally, its cost is treated as zero. Moreover, the target-side prefill and decode cost is evaluated using a proxy price of $\$0.80$ per million input tokens and $\$0.80$ per million output tokens, selected to align with market pricing. 


\begin{table*}[t]
\centering
\small
\setlength{\tabcolsep}{4.0pt}
\renewcommand{\arraystretch}{1.12}
\begin{tabular}{@{}p{0.2\textwidth}cccccccccc@{}}
\toprule
\multirow{2}{*}{Variant}
& \multicolumn{6}{c}{Enabled components}
& \multicolumn{4}{c}{Performance} \\
\cmidrule(lr){2-7}
\cmidrule(lr){8-11}
& Vis. Red. & Tok. Sel. & M-Gate & Len. Adapt. & Branch & DVC
& Spd. $\uparrow$ & Acc. $\uparrow$ & Comm. $\downarrow$ & Cost Red. $\uparrow$ \\
\midrule
Full-token drafting
& -- & -- & $\checkmark$ & $\checkmark$ & $\checkmark$ & $\checkmark$
& 1.64$\times$ & 68.16 & \underline{14.32} & \textbf{34.81} \\

Random-token selection
& $\checkmark$ & -- & $\checkmark$ & $\checkmark$ & $\checkmark$ & $\checkmark$
& 1.71$\times$ & 58.57 & \textbf{13.59} & \underline{34.65} \\
\midrule
w/o Margin gating
& $\checkmark$ & $\checkmark$ & -- & $\checkmark$ & $\checkmark$ & $\checkmark$
& 1.73$\times$ & \textbf{74.70} & 15.63 & -6.90 \\

w/o Adaptive drafting 
& $\checkmark$ & $\checkmark$ & $\checkmark$ & -- & $\checkmark$ & $\checkmark$
& \underline{2.14$\times$} &71.88 & 16.94  &30.28\\

w/o Parallel branching
& $\checkmark$ & $\checkmark$ & $\checkmark$ & $\checkmark$ & -- & $\checkmark$
& 2.07$\times$ & \underline{74.40} & 16.49 & 29.75 \\

w/o Decoupled correction
& $\checkmark$ & $\checkmark$ & $\checkmark$ & $\checkmark$ & $\checkmark$ & --
& 1.98$\times$  &\underline{74.40}& 82.68 & 29.75 \\

\rowcolor{proposedblue}
\textbf{CoVSpec}
& $\checkmark$ & $\checkmark$ & $\checkmark$ & $\checkmark$ & $\checkmark$ & $\checkmark$
& \textbf{2.21}$\times$ & \underline{74.40}& 16.49 & 29.75\\
\bottomrule
\end{tabular}
\caption{Component ablation of CoVSpec on VQAv2 at $\mathrm{SNR}=10$ dB. Vis. Red. indicates visual-token reduction for device-side drafting, and Tok. Sel. indicates the proposed representative visual token selection. M-Gate, Len. Adapt., Branch, and DVC denote margin-based gating, adaptive draft-length control, parallel branching, and decoupled verification-correction, respectively.}
\label{tab:ablation_all}
\end{table*}
\subsection{Main Results}
Table~\ref{tab:main_results} compares the proposed CoVSpec with Device-only, Edge-only, Vanilla SD~\cite{leviathan2023fast}, and U-HLM~\cite{JP_hybrid} at $\mathrm{SNR}=10$ dB on VQAv2, MMMU, and MMBench. Specifically, Device-only runs the small VLM on the mobile device, Edge-only runs the large VLM solely on the edge server, and Vanilla SD denotes
  the standard device--edge speculative decoding. For CoVSpec, 64 out of 768 visual tokens are retained for mobile-side drafting. From this table, we can observe that CoVSpec achieves the best overall trade-off among decoding speed, communication cost, inference cost, and task accuracy. Specifically, compared with Edge-only, CoVSpec achieves a speedup of $2.21\times$, $1.63\times$, and $1.85\times$ on VQAv2, MMMU, and MMBench, respectively, while reducing the inference cost to $0.71\times$, $0.85\times$, and $0.77\times$. Moreover, compared with Vanilla SD, CoVSpec reduces the communication cost from hundreds of MB to less than 22 MB across all benchmarks. In contrast to Device-only, CoVSpec attains substantially higher answer accuracy by leveraging the edge-server large VLM for verification. These results demonstrate that CoVSpec significantly improves end-to-end efficiency while maintaining competitive task performance in device-edge VLM co-inference.

\subsection{Ablation Studies}

In Table~\ref{tab:ablation_all}, we conduct ablation experiments to verify the effectiveness of the key components of CoVSpec on VQAv2 at $\mathrm{SNR}=10$ dB. From this table, we can observe that representative visual token selection improves both efficiency and accuracy: compared with full-token drafting, it increases the speedup from 1.64$\times$ to 2.21$\times$ and the accuracy from 68.16\% to 74.40\%, since too many visual tokens confuse the small draft model and lead to wrong but confident predictions that skip target verification. Compared with random-token selection, it further raises the speedup from 1.71$\times$ to 2.21$\times$ and the accuracy from 58.57\% to 74.40\%, as random tokens give the draft model little useful visual information, so it relies on language priors and produces confident but unreliable outputs that also bypass target correction. In addition, removing margin gating lowers the speedup from 2.21$\times$ to 1.73$\times$ and changes the cost reduction from 29.75\% to $-6.90$\%, confirming its role in avoiding unnecessary device--edge synchronization, while removing adaptive draft-length control still yields a 2.14$\times$ speedup, suggesting that it mainly improves communication-awareness. Finally, parallel branching improves speed by overlapping drafting with verification, while decoupled verification-correction does not affect accuracy and reduces communication from 82.68 MB to 16.49 MB by moving residual sampling to the device, improving the speedup from 1.98$\times$ to 2.21$\times$. These results further demonstrate the effectiveness of the proposed CoVSpec framework.

\section{Conclusion}
In this paper, we have proposed CoVSpec, an efficient device--edge co-inference framework for VLM inference. By combining representative visual token selection, adaptive drafting, and decoupled verification-correction, CoVSpec reduces mobile-side drafting cost and device--edge communication overhead. Experiments on multiple benchmarks demonstrated that CoVSpec achieves up to $2.21\times$ higher decoding speed than the target-only baseline and reduces communication overhead by over $96\%$, while maintaining competitive answer accuracy.

\bibliographystyle{IEEEtran}
\bibliography{IEEEabrv,references}
\end{document}